\documentclass[10pt,twocolumn,letterpaper]{article}

\usepackage[pagenumbers]{wacv} 

\usepackage{graphicx}
\usepackage{amsmath}
\usepackage{amssymb}
\usepackage{booktabs}
\usepackage[accsupp]{axessibility}

\usepackage{xcolor}
\usepackage{bbm}
\usepackage{pifont}
\usepackage{esvect}
\usepackage{rotating}
\definecolor{fgreen}{rgb}{0.13, 0.55, 0.13}
\definecolor{bred}{rgb}{0.55, 0.13, 0.13}

\DeclareMathOperator*{\argmax}{argmax}

\newcommand{\lturn}[1]{\begin{turn}{90} #1 \end{turn}}

\usepackage[pagebackref,breaklinks,colorlinks]{hyperref}

\usepackage[capitalize]{cleveref}
\crefname{section}{Sec.}{Secs.}
\Crefname{section}{Section}{Sections}
\Crefname{table}{Table}{Tables}
\crefname{table}{Tab.}{Tabs.}

\def\wacvPaperID{202} 
\def\confName{WACV}
\def\confYear{2024}

\begin{document}

\title{2D Feature Distillation for Weakly- and Semi-Supervised \\ 3D Semantic Segmentation}

\author{
Ozan Unal$^1$ \quad Dengxin Dai$^{2}$ \quad Lukas Hoyer$^1$ \quad Yigit Baran Can$^1$ \quad   Luc Van Gool$^{1,3,4}$ \\
$^1$ETH Zurich, $^2$Huawei Technologies, $^3$KU Leuven, $^4$INSAIT \\
{\tt\small \{ozan.unal, dai, lukas.hoyer, cany, vangool\}@vision.ee.ethz.ch}
}

\maketitle

\begin{abstract}
As 3D perception problems grow in popularity and the need for large-scale labeled datasets for LiDAR semantic segmentation increase, new methods arise that aim to reduce the necessity for dense annotations by employing weakly-supervised training. However these methods continue to show weak boundary estimation and high false negative rates for small objects and distant sparse regions. We argue that such weaknesses can be compensated by using RGB images which provide a denser representation of the scene. We propose an image-guidance network (IGNet) which builds upon the idea of distilling high level feature information from a domain adapted synthetically trained 2D semantic segmentation network. We further utilize a one-way contrastive learning scheme alongside a novel mixing strategy called FOVMix, to combat the horizontal field-of-view mismatch between the two sensors and enhance the effects of image guidance. IGNet achieves state-of-the-art results for weakly-supervised LiDAR semantic segmentation on ScribbleKITTI, boasting up to $98\%$ relative performance to fully supervised training with only $8\%$ labeled points, while introducing no additional annotation burden or computational/memory cost during inference. Furthermore, we show that our contributions also prove effective for semi-supervised training, where IGNet claims state-of-the-art results on both ScribbleKITTI and SemanticKITTI. 
\end{abstract}

\vspace{-12px} \section{Introduction}

With the ever growing interest in 3D scene understanding for autonomous vehicles, semantic segmentation for LiDAR point clouds has also risen in popularity. To accurately and robustly learn the dense prediction task of generating per point class labels, a high volume of data is not only valuable but required. However manually labeling outdoor LiDAR scenes for semantic segmentation is both time consuming and expensive for large scale datasets.

There are two recently explored paths in the literature for reducing the labeling cost of outdoor LiDAR scenes: (i) by employing weak-supervision, where all frames have incomplete labels (e.g. by using line-scribbles~\cite{Unal_2022_CVPR}) and (ii) by employing semi-supervision, where a subset of frames are labeled and the rest remain completely unlabeled~\cite{iccv2021guided}. 

\begin{figure}
    \centering
    \includegraphics[width=\columnwidth]{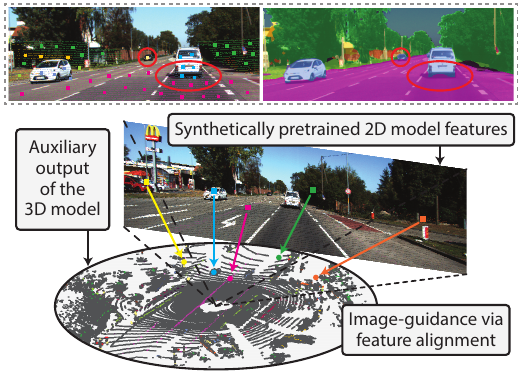}
    \caption{While boundaries and sparse distant regions are difficult to determine in 3D, 2D models can leverage the denser image pixels for finer estimation. With image-guidance via feature alignment, points with pixel correspondences aim to mimic the 2D model features via an auxiliary loss.}
    \label{fig:overview}
\vspace{-12px} \end{figure}

Commonly, LiDAR semantic segmentation models suffer from error prone boundary estimation between classes, as well as high false negative rates on both small objects and distant sparse regions. This is caused by the sparsity of LiDAR point clouds which severely reduces the number of points that fall on such regions to form an understandable and well separable geometry. As expected, these errors are further amplified when dealing with incomplete supervision, especially with scribble labels that completely forgo labeling boundaries. It can even be argued that such hard cases potentially need more representation within the dataset for correct and robust learning, something that clearly lacks under data-efficient settings.

These errors are severely reduced when operating on a denser representation of a scene (see Fig.~\ref{fig:overview} - top). Luckily, LiDAR sensors are commonly paired with cameras that are not only cheaper but also provide a dense signal in the form of an RGB image that allows better separable boundaries (especially with the aid of RGB color channels), as well as orders of magnitude more pixels than points on small objects and distant regions. It is for this reason that all autonomous vehicles are equipped with a high resolution camera facing the front of the car to provide a denser and more complete understanding of the critical ego-vehicle path.

Our goal in this work is to leverage this high resolution image within our 3D pipeline to target the common weaknesses of LiDAR semantic segmentation models trained under incomplete supervision (weak labels). However we face two major challenges: (i) we need to retain our low annotation budget to have a scalable solution, therefore we cannot use additional annotated datasets or pretrained models in our setup; (ii) we need to tackle the issue of the horizontal field-of-view (FOV) mismatch between a LiDAR sensor and camera, where only a subset of points that fall onto the camera FOV have valid correspondence.

To this extent, we propose the Image-Guidance network (IGNet) that comprises of two core modules: (\textbf{M1}) domain adaptive image feature distillation that allows us to keep our low annotation budget and (\textbf{M2}) one-way contrastive learning that combats the FOV mismatch by leveraging image features to supervise out-of-image points. Throughout this work, we strictly associate the 2D domain with RGB images and 3D with LiDAR point clouds.

\noindent \textbf{M1}: Firstly, we train a 2D semantic segmentation model to generate per pixel high level features that better capture shape and context for sparse regions. By training on synthetic data, we avoid introducing any additional annotation requirements. We establish point-to-pixel correspondence between the LiDAR point cloud and the camera image (Fig.~\ref{fig:overview} - bottom), and distill the information from the generated features onto a 3D network via an auxiliary loss.

However, training on synthetic data yields yet another challenge: There exists a domain gap between synthetic images and real images that hinder performance in 2D. To further improve the quality of our image features, we propose using a domain adaptation (DA) pipeline to align our source domain onto the target. We further supervise the DA task via weak image labels generated by projecting the LiDAR labels onto the corresponding image.

\noindent \textbf{M2:} Next, we tackle the issue of the horizontal FOV mismatch between the camera and the LiDAR sensor. As our image-guidance module requires valid point-pixel correspondences, the auxiliary supervision remains limited to points that fall onto the image. To extend the supervision to points outside of the image, we propose using a one-way contrastive loss guided by a teacher model, allowing points that fall within the image to guide points that fall outside.

Here we observe that the number of pixel-to-outside-point-pairings remains limited as each LiDAR scan has a fixed associated image. This reduces the effect of the contrastive learning, especially since this single image alone often contains zero to a few object instances of each class. To combat this, we introduce a simple mixing strategy called FOVMix, where we cut and paste an image with its corresponding points from one scene onto another. With FOVMix, we are not only able to generate new pixel-point pairings to aid the contrastive learning but also increase the variability within each mini-batches.

\noindent To summarize:
\begin{itemize}
    \setlength\itemsep{-0.1em}
    \item We propose using a synthetically trained 2D semantic segmentation model to guide the 3D network's feature space in order to improve boundary, distant region and sparse object segmentation.
    \item We employ weakly-supervised domain adaptation to further align the 2D features with our dataset.
    \item We extend the supervision from the image-guidance network to points out of the camera field-of-view via a one-way supervised contrastive loss.
    \item We propose a new mixing strategy called FOVMix to introduce additional variety into the dataset along with additional point-pixel pairings to extract further performance from our contrastive loss.
\end{itemize}
We achieve state-of-the-art results for weakly-supervised semantic segmentation on ScribbleKITTI~\cite{Unal_2022_CVPR}. We further show that IGNet can also be utilized for semi-supervised LiDAR segmentation to yield state-of-the-art results on both ScribbleKITTI and SemanticKITTI~\cite{iccv2019semantickitti}.

It should be noted that our proposed modules are only required during training, thus the performance boost comes without any additional computational or memory burden compared to the baseline 3D model during inference. Finally, as only synthetic data is required, we also do not introduce any additional annotation costs.

\section{Related Work}
\noindent \textbf{Data Efficient LiDAR Semantic Segmentation:} LiDAR semantic segmentation research has heavily focused on understanding how to best process the unordered data structure, with earlier focus on direct point based neural networks~\cite{cvpr2017pointnet, arxiv2017pointnet++, wacv2021improving, cvpr2020randla,iccv2019kpconv} having later shifted to sparse convolutional networks~\cite{cvpr2021cylindrical, eccv2020spvnas, cvpr2019minkowski, arxiv2020amvnet, arxiv2020sparse}. As architectures mature, we observe another developing area of interest: data efficiency within LiDAR semantic segmentation.

As known, the dense prediction task requires a large-scale annotated dataset, which is especially difficult and expensive to obtain for LiDAR point clouds~\cite{iccv2019semantickitti}. Recent work therefore investigate two paths that aim to reduce this associated labeling cost: (i) weakly-supervised learning, where every frame is partially labeled, and (ii) semi-supervised learning, where only a subset of frames are labeled and the remaining stay completely unlabeled. However such approaches always come at the cost of performance, as reducing the number of labels within a dataset reduces the supervision provided to the model. Current popular literary work that deal with incomplete labels aim to extend the supervision to unlabeled points by (i) self-supervised training~\cite{aaai2020curriculum, eccv2018classbalanced, iccv2021dars} where a model is trained on self-generated pseudo-labels or (ii) relying on a guidance network to generate on the fly targets (e.g. mean teacher~\cite{nips2017meanteacher, Unal_2022_CVPR, unal2023discwise}).

For self-supervised training, CBST~\cite{eccv2018classbalanced} proposes to use class-wise thresholding for self-training to reduce confirmation bias. Extending CBST, DARS~\cite{iccv2021dars} proposes to re-distribute biased pseudo labels for semi-supervised training.

For 3D in particular, ScribbleKITTI~\cite{Unal_2022_CVPR} provides the first realistic benchmark for weakly supervised LiDAR semantic segmentation by introducing the scribble-annotated dataset. In their work, to reduce the gap to fully supervised training, they propose the SSLSS pipeline where they utilize a mean teacher setup~\cite{nips2017meanteacher} to stretch the supervision to unlabeled points, and extend CBST with a range component to deal with the increased sparsity of LiDAR point clouds. 
For works on indoor point clouds, PSD~\cite{eccv2020fusionnet} utilizes similar consistency checks to align clean and perturbed outputs of unlabeled points. WS3D~\cite{liu2022weakly} utilizes region-level boundary awareness and instance discrimination to improve indoor and outdoor 3D semantic segmentation with simulated weak labels. Furthermore for semi-supervised learning, DiAL~\cite{unal2023discwise} uses a simple MT setup, GPC~\cite{iccv2021guided} proposes using a pseudo-label guided point contrastive loss, SSPC~\cite{arxiv2021sspc} utilizes self-training and LaserMix~\cite{kong2022lasermix} uses a mixing operation to bring supervision to unlabeled frames. CPS~\cite{chen2021semi} utilizes a Siamese structure to induce cross supervision.

\noindent \textbf{Multi-Modality with LiDAR and Image:} As mentioned, the additional information available in the corresponding RGB image does provide meaningful advantages that can improve LiDAR perception. Yet the task of incorporating this information within a robust pipeline is not trivial.

Fusion has been studied for a number of LiDAR based 3D perception tasks in a supervised and weakly-supervised manner ~\cite{caltagirone2019lidar, bai2022transfusion, li2022deepfusion, 8760386, zhong2021survey, meng2020weakly}. For LiDAR semantic segmentation PMF~\cite{Zhuang_2021_ICCV} and LIF-Seg~\cite{zhao2021lif} fuse the information from streams that process each modality individually to obtain higher information yielding features. However such approaches not only require image information during inference but also have linearly increasing memory and computation cost. 2DPASS~\cite{yan20222dpass} overcomes this by only using a one way information flow during training. Still, training the image stream on only LiDAR projected labels suffer heavily under incomplete annotations where it hinders performance instead of improving it. Sautier~\etal~\cite{sautier2022image} proposes a more general approach of self-supervised pretraining through the alignment of pixel- and point regions that still remains susceptible to forgetting (at a reduced scale).

\noindent \textbf{Mix-Augmentation:} Mixing operations have been very successful in increasing variability in the dataset and producing significant performance boosts for many tasks~\cite{yun2019cutmix, chen2022stackmix, zhang2022cyclemix, 9240708, olsson2021classmix, franchi2021robust, hoyer2021three}. CutMix~\cite{yun2019cutmix} mixes portions of the input and output of one sample image with another. MixMatch~\cite{nips2019mixmatch} applies the same mixing operation to labeled and unlabeled frames in a semi-supervised setting while generating labels via guessing and sharpening for unlabeled parts to provide supervision. Specifically for semi-supervised learning on LiDAR point clouds, LaserMix~\cite{kong2022lasermix} aims to introduce variability through cylindrical and range-view partitioning and mixing.

\section{Data Efficient LiDAR Segmentation}

Data efficient LiDAR semantic segmentation aims to reduce the labeling cost associated with the dense prediction task by employing (i) weak supervision, where all frames have incomplete labels (e.g. by using scribble annotations), or (ii) semi supervision, where some frames have labels and others remain unlabeled. In either setting, naively training a model on available labeled points results in a considerable performance drop as only a small subset of points provide supervision. Specifically, we observe an amplified error rate caused by (i) weak boundary estimation between classes and (ii) misclassification of small objects and distant sparse regions, as LiDAR's increased sparsity by range causes a severe reduction in the number of available points on an object to form an understandable geometry.

\subsection{A Baseline Approach: Mean Teacher}

\begin{figure*}[t]
    \centering
    \includegraphics[width=\textwidth]{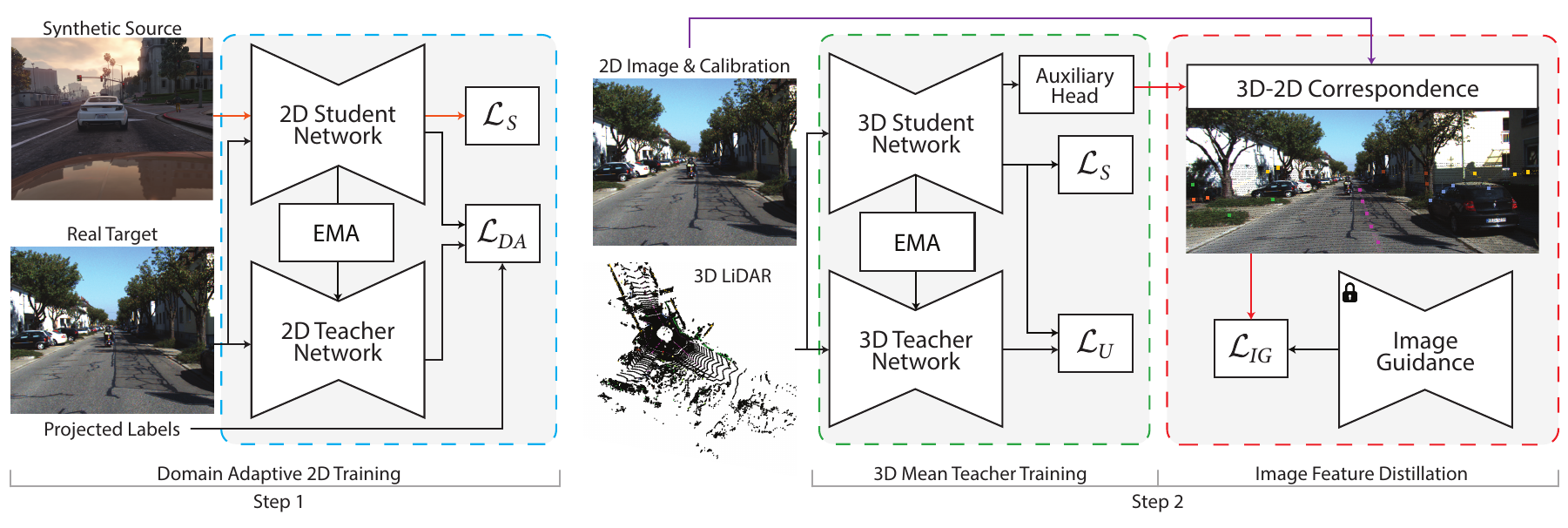}
    \caption{Pipeline for image feature distillation. We first establish point-pixel correspondences between the LiDAR point cloud and image. (blue) The available weak point labels are then used to generate weak image labels that supervise a 2D network alongside synthetic data. We utilize a mean teacher framework to adapt from the synthetic domain to the real domain. (green) We train a 3D model using a mean teacher framework to utilize both weak annotations and unlabeled points. (red) We copy and freeze the trained 2D student model to generate per pixel features that act as a guidance for the 3D student features via an auxiliary loss.}
    \label{fig:pipeline}
\vspace{-12px}\end{figure*}

As a first step in reducing the performance gap to fully supervised training we employ a generalized approach to utilize all points within the dataset. In specific, to extend the supervision to unlabeled points, following Unal~\etal~\cite{Unal_2022_CVPR}, we construct a mean teacher (MT) framework~\cite{nips2017meanteacher}, where a student network is trained using a supervised loss $H$ (e.g. cross-entropy) and a teacher network is formed by the exponential moving average (EMA) of the student's weights $\theta$ (for time step $t$):
\begin{equation} \label{eq:ema}
    \theta^\textrm{EMA}_t = \alpha \theta^\textrm{EMA}_{t-1} + (1-\alpha) \theta_t
\end{equation}
The given update rule yields a teacher model that is a better and more robust predictor~\cite{siam1992averaging, nips2017meanteacher}. To exploit this behaviour, we apply a consistency loss between the teacher and the student to align its outputs to the more accurate predictions, e.g. by minimizing the Kullback-Leibler divergence to the softmax outputs. Formally, for all points $x$, the loss function can be redefined as:
\begin{equation} \label{eq:partial_consistency}
    \mathcal{L} = H(\hat{\mathrm{y}}, y) + \mathbbm{1}_U(x) \, \textrm{KL}(\hat{\mathrm{y}} \, || \, \hat{\mathrm{y}}_\textrm{EMA})
\end{equation}
with $\hat{\mathrm{y}}$ and $\hat{\mathrm{y}}_\textrm{EMA}$ denoting the predictions of the student and teacher models, $y$ the ground truth labels and $U$ denoting the set of points without ground truth labels. An illustration of the MT pipeline can be seen in Fig.~\ref{fig:pipeline} - green.

While a mean teacher framework does allow us to utilize the entire dataset within our training pipeline, due to the lack of direct supervision, similar to the student, the teacher's predictions remain uncertain and error prone for points that lie on class boundaries or for sparsely represented classes (e.g. volumetrically small objects or distant regions), especially when trained on weak scribble labels that completely forgo labeling any boundary points.

\subsection{Image Guidance via Feature Distillation}

To target these weaknesses we propose using image feature distillation from a trained 2D semantic segmentation model. But before we dive deep into the details, it is important to establish motivation.

RGB images provide a much denser representation of a scene compared to LiDAR point clouds. This increased density along with the available color channels allow easier distinction of both class boundaries as well as small objects and distant regions. 2D semantic segmentation models can therefore learn better separable and richer features for such pixels. Following this observation, we propose introducing an image guidance (IG) network to exploit the mature features of a trained 2D semantic segmentation model.

Firstly, we apply a forward pass to the camera image using a synthetically-trained semantic segmentation model to extract a high level feature representation ($\theta_\textrm{IG}:[0,255]^3 \mapsto f_{IG} \in \mathbb{R}^d$). It should be noted that we opt to use synthetic data to avoid introducing any additional annotation burden as the collection of new labeled samples can be easily automated. Using available intrinsic and extrinsic camera matrices $K$ and $[R | t]$ respectively, we project the 3D points cloud in homogeneous coordinates $x_{hom}$ onto the rectified camera coordinates following $x^T_{rec} = K [R | t]  \, x^T_{hom}$
and extract point to pixel mappings $ m: x_{rec} \mapsto (k,l)$
with $k = \lfloor x_{rec}^{(0)} /x_{rec}^{(2)} \rfloor$ and $\lfloor l = x_{rec}^{(1)} /x_{rec}^{(2)}\rfloor$.
A point to pixel correspondence is considered valid if the pixel $(k,l)$ falls within the image.

We extend our 3D model with an auxiliary head that maps the final layer features to the image feature dimension $d$. During training, we introduce a new consistency term between the student and the IG teacher that is applied to all points that have a valid pixel correspondence. Formally, we restate the loss function to include image-guidance as:
\begin{equation} \label{eq:image_guidance}
\begin{split}
    \mathcal{L} = & H(\hat{\mathrm{y}}, y) + \mathbbm{1}_U(x) \, \textrm{KL}(\hat{\mathrm{y}} \, || \, \hat{\mathrm{y}}_\textrm{EMA})
    + \mathcal{L}_{IG} \\
    & \textrm{with } \mathcal{L}_{IG} = \mathbbm{1}_I(x, m(x)) \, \textrm{KL}(sm(f) \, || \, sm(f_\textrm{IG}))
\end{split}
\end{equation}
with $I$ denoting the set of points with valid pixel correspondence, $sm$ denoting the softmax operation, $f$, $f_{IG} \in \mathbbm{R}^{N' \times C}$ denoting the feature representations of the 3D auxiliary head and IG decoders respectively.

With the addition of the auxiliary loss, the 3D network aims to mimic the more mature representation of the 2D network for points with pixel correspondences. In other words, we introduce a new teacher model, where boundary points along with small and distant objects more richly defined due to the denser representation, to further and better guide the student on unlabeled points. An illustration of the proposed module can be seen in Fig.~\ref{fig:pipeline} - red.

It should be noted that the IG network is only required during training and can be completely removed for inference alongside the auxiliary head, causing no additional memory requirements or time costs to the overall 3D model.

\subsection{2D Weakly-Supervised Domain Adaption}

As mentioned before, in order to train $\theta_{IG}$ for semantic segmentation, we resort to synthetic data. It has the desirable property that even dense annotations can be automatically generated so that no additional labeling cost is introduced. However, a model trained on synthetic source data $(I_s,S_s)$, usually experiences a performance drop when applied to real-world target images $I_t$ due to the domain gap.

To tackle this, we propose employing a domain adaptation pipeline to improve the quality of the extracted features and better align with the data from our real-world training set. Following current literature~\cite{hoyer2021daformer}, we reestablish a mean teacher framework~\cite{nips2017meanteacher} and use the teacher model to generate pseudo labels $P_t$ for the target domain images by freezing the unlabeled image predictions. We train the 2D network with a linear classification layer $\gamma$ not only on the synthetic image-label pairings ($I_s$, $S_s$) but also on the target images with pseudo labels ($I_t$, $P_t$). Formally, the loss for the 2D model can be defined as:
\begin{equation} \label{eq:domain_adaptation_loss}
\begin{split}
    & \textrm{\phantom{with}} \mathcal{L} = \mathcal{L}_{S} +  \mathcal{L}_{DA} \\
    & \textrm{with }  \mathcal{L}_{S} = H(\gamma(\theta_\mathit{IG}(I_s)), S_s) \\
    & \textrm{and }  \mathcal{L}_{DA} =  H(\gamma(\theta_\mathit{IG}(I_t)), P_t)
\end{split}
\end{equation}

Furthermore, in contrast to common unsupervised domain adaptation, we have access to LiDAR scribble annotations on the target domain. Even though these only provide sparse and possibly noisy supervision  (due to projection errors), they can be an important anchor for the adaptation to the target domain. In order to incorporate this additional information into our pipeline, we augment the EMA teacher pseudo-label $P_t$ with projected scribble labels $P_t(m(x)) \leftarrow y$.

We then extend our domain adaptive loss $\mathcal{L}_{DA}$ from Eq.~\ref{eq:domain_adaptation_loss} to increase the importance of the projected labels $P_t(m(x))$ via a weight vector $\vv{\lambda_p}$:
\begin{equation}
    \mathcal{L}_{DA} = \vv{\lambda_p} H(\gamma(\theta_\mathit{IG}(I_t)), P_t)
\end{equation}
with $\vv{\lambda_p} = \lambda_p$ for pixels with valid point mapping and 1 otherwise.
An illustration of the proposed weakly-supervised domain adaptation pipeline can be seen in Fig.~\ref{fig:pipeline} - blue.

Finally, to form the image guidance model $\theta_{IG}$, we copy and freeze the 2D student model (following unsupervised domain adaptation convention~\cite{hoyer2021daformer}) without the linear classifier and use its generated features to guide the 3D student model during training.

\subsection{Extending the Supervision Beyond the Image}

\begin{figure}
    \centering
    \includegraphics[width=\columnwidth]{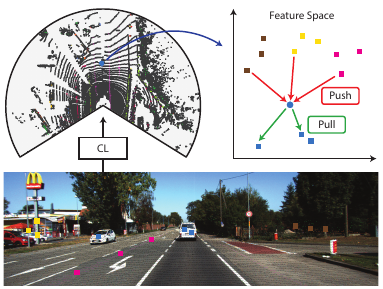}
    \caption{Illustration of the one-way supervised contrastive loss. Points with pixel correspondence guide points outside of the image field-of-view via pull and push forces applied based on available weak labels.}
    \label{fig:1cl}
\vspace{-12px}\end{figure}

With image-guidance (Eq.~\ref{eq:image_guidance}) the information distillation from the mature 2D features to the 3D pipeline is limited by the availability of point-pixel correspondences. For many cases, we are limited to a front facing camera, so there exists a big mismatch between the horizontal FOV of the two sensors. Under such a setup, the set of all points with valid pixel correspondence ($I$) is much smaller than the set of all points without a valid correspondence ($O = I \cap P$), i.e. $|I|<|O|$. In other words, the lack $360^{\circ}$ coverage for the camera means that points with pixel correspondence only make up a small portion of the LiDAR point cloud.

To be able to guide points outside of the image using the 2D domain adapted features, we introduce an extension to the image-guidance loss with a one-way supervised contrastive loss (CL).

Let $I^{(c)} \subseteq I$ and $O^{(c)} \subseteq O$ define two sets of points inside and outside of the image respectively with associated class $c = \argmax \, \hat{\mathrm{y}}_\textrm{EMA}$, given by the teacher's prediction. Formally, we define the one-way supervised contrastive loss as:
\begin{equation}
    \mathcal{L}_{CL} = \! \sum_c \! \!  \! \sum_{o\in O^{(c)}} \! \! \! - \! \log \! \left( \! 
    \frac{1}{|O^{(c)}|} \sum_{i \in I^{(c)}}  \! \frac{\exp (f_o \cdot f_{IG,i} / \tau)}{\sum\limits_{i' \in I} \exp (f_o \cdot f_{IG,i'} / \tau)}
    \! \right)
\end{equation}
with $\tau$ denoting the temperature. The total loss can then be formulated as:
\begin{equation} \label{eq:total_loss}
\begin{split}
    \mathcal{L} = & H(\hat{\mathrm{y}}, y) + \mathbbm{1}_U(x) \, \textrm{KL}(\hat{\mathrm{y}} \, || \, \hat{\mathrm{y}}_\textrm{EMA})
    + \mathcal{L}_{IG} + \lambda \mathcal{L}_{CL}
\end{split}
\end{equation}
with $\lambda$ denoting the scale hyperparameter.

As illustrated in Fig.~\ref{fig:1cl}, the loss extension aims to apply a pull force to all points towards pixels of the same category while also applying a push to all points away from pixels of a different class. We therefore align the features of points outside of the image with the features of the 2D image-guidance network.

\begin{figure}
    \centering
    \includegraphics[width=\columnwidth]{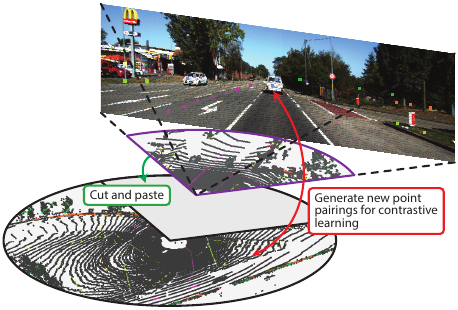}
    \caption{Illustration of the proposed mixing strategy FOVMix that not only increases the variety within the training set but also generates new point pairing inside-outside of the image field-of-view to further guide all points.}
    \label{fig:fovmix}
\vspace{-12px}\end{figure}

\subsection{FOVMix}

Finally, we introduce a new mixing operation called FOVMix. Given two data samples $(x_A, y_A, I_A)$ and $(x_B, y_B, I_B)$, the goal of FOVMix is to generate a
new training sample $(\tilde{x}, \tilde{y}, \tilde{I})$. Simply put, we take an image from sample A and replace it with the image of sample B. To accompany this, we further take all points that are within the image FOV of sample A, and paste them onto sample B while removing all points of B that were in the same region. An illustration of FOVMix can be seen in Fig.~\ref{fig:fovmix}.

\noindent Formally, we define the mixing operation as:
\begin{equation}
\begin{split}
    \tilde{x} &= [\mathbf{M}_{AA} \odot x_A, (1 - \mathbf{M}_{AA}) \odot x_B] \\
    \tilde{y} &= [\mathbf{M}_{AB} \odot y_A, (1 - \mathbf{M}_{AB}) \odot y_B] \\
    \tilde{I} &= I_A
\end{split}
\end{equation}
$\mathbf{M}_{AB}$, $\mathbf{M}_{AA} \in \{0,1\}^N$ denote the binary masks that yield the points within the image FOV given the intrinsic projection matrix $A$ and extrinsic projection matrices $A$ and $B$ respectively, $\odot$ and [,] denoting a dot product for masking and concatenation operations. Thus, FOVMix does not depend a specific sensor/setting, but only relies on the availability of point to pixel correspondences, which is expected for systems with both a LiDAR sensor and camera.

FOVMix is a simple operation that accomplishes two feats: (i) it increases the effectiveness of the one-way contrastive loss by introducing additional pairings of points inside-outside of the image, (ii) it increases the richness of the data within each mini-batch. While FOVMix introduces noise along the boundaries of the image FOV similar to other mixing methods commonly used in dense vision tasks, the increased diversity and richness of each mini-batch is a worthy trade-off against the introduced noise.

\section{Experiments}


\noindent \textbf{Implementation details:} We use Cylinder3D~\cite{cvpr2021cylindrical} as a baseline 3D model. For the mean teacher, we follow convention and set the update hyperparameter $\alpha = 0.999$~\cite{unal2023discwise}. For the domain adaptive 2D pipeline we follow DAFormer~\cite{hoyer2021daformer}. We heuristically balance the losses by setting $\lambda=0.001$ and $\lambda_p=10$. For semi-supervised, we restrict set $A$ in FOVMix to labeled frames to ensure we have direct supervision in all samples and do additional rotation augmentation before the FOVMix operations to increase variability.

\noindent \textbf{Datasets:} We run our experiments on the ScribbleKITTI~\cite{Unal_2022_CVPR} dataset that provides realistic weak labels for LiDAR semantic segmentation in the form of scribbles. ScribbleKITTI is built on SemanticKITTI~\cite{iccv2019semantickitti, ijrr2013kitti}, the most popular large-scale outdoor-scene dataset for LiDAR semantic segmentation, shares the same \textit{valid}-set. The weak labels only provide annotations to $8\%$ of the point count and completely forgo class boundaries. Thus, compared to dense annotations, labeling times are reduced by 10 fold.

For the 2D syntetic training, we use the GTA-V dataset which contains 24966 synthetic images with pixel level semantic annotation. The images are generated using a modded version of the open-world video game Grand Theft Auto 5.



\subsection{Results}

\begin{table*}[t]
    \tabcolsep=0.11cm
    \resizebox{\textwidth}{!}{
    \begin{tabular}{|l|c|ccccccccccccccccccc|}
        \hline
        Method
        & mIoU
        &\lturn{car}
        &\lturn{bicycle}
        &\lturn{m.cycle}
        &\lturn{truck} 
        &\lturn{o.vehicle } 
        &\lturn{person}
        &\lturn{bicyclist} 
        &\lturn{m.cyclist} 
        &\lturn{road}
        &\lturn{parking}  
        &\lturn{sidewalk} 
        &\lturn{o.ground} 
        &\lturn{building} 
        &\lturn{fence}
        &\lturn{vegetation } 
        &\lturn{trunk}
        &\lturn{terrain} 
        &\lturn{pole}
        &\lturn{t.sign} \\
        [0.5ex] 
        \hline
        Cylinder3D~\cite{cvpr2021cylindrical} & 57.0 & 88.5 & 39.9 & 58.0 & 58.4 & 48.1 & 68.6 & 77.0 & \textbf{0.5} & 84.4 & 30.4 & 72.2 & 2.5 & 89.4 & 48.4 & 81.9 & 64.6 & 59.8 & 61.2 & 48.7 \\
        MinkNet*~\cite{cvpr2019minkowski} & 58.5 & 91.1 & 23.8 & 59.0 & 66.3 & 58.6 & 65.2 & 75.2 & 0.0 & 83.8 & 36.1 & 72.4 & 0.7 & 90.2 & 51.8 & \textbf{86.7} & 68.5 & \textbf{72.5} & 62.5 & 46.6 \\
        SPVCNN*~\cite{eccv2020spvnas} & 56.9 & 88.6 & 25.7 & 55.9 & 67.4 & 48.8 & 65.0 & 78.2 & 0.0 & 82.6 & 30.4 & 70.1 & 0.3 & 90.5 & 49.6 & 84.4 & 67.6 & 66.1 & 61.6 & 48.7 \\
        MT~\cite{nips2017meanteacher} & 59.0 & 91.0 & 41.1 & 58.1 & \textbf{85.5} & 57.1 & 71.7 & 80.9 & 0.0 & 87.2 & 35.1 & 74.6 & 3.3 & 88.8 & 51.5 & 86.3 & 68.0 & 70.7 & 63.4 & 49.5 \\
        CBST~\cite{eccv2018classbalanced} & 60.8 & \textbf{92.4} & 39.1 & 58.5 & 78.5 & 57.0 & 70.0 & 77.4 & 0.0 & 86.9 & 35.4 & 74.3 & 7.3 & 89.8 & \textbf{55.6} & 85.1 & 66.7 & 68.1 & 62.0 & 51.1 \\
        DARS~\cite{iccv2021dars} & 60.8 & 91.9 & 39.3 & 57.9 & 78.6 & 53.3 & 69.5 & 77.1 & 0.0 & 86.6 & \textbf{37.2} & 74.2 & \textbf{8.3} & 89.8 & 54.5 & 86.5 & 68.8 & 70.1 & 63.4 & 49.0 \\
        SSLSS~\cite{Unal_2022_CVPR} & 61.3 & 91.0 & 41.1 & 58.1 & \textbf{85.5} & 57.1 & 71.7 & 80.9 & 0.0 & 87.2 & 35.1 & 74.6 & 3.3 & 88.8 & 51.5 & 86.3 & 68.0 & 70.7 & 63.4 & 49.5 \\
        \hline
        IGNet (Ours) & \textbf{62.0} & 90.7 & \textbf{47.6} & \textbf{64.5} & 83.2 & \textbf{60.5} & \textbf{74.5} & \textbf{81.3} & 0.0 & \textbf{88.6} & 34.6 & \textbf{75.5} & 2.3 & \textbf{90.6} & 53.0 & 83.5 & \textbf{69.5} & 63.7 & \textbf{63.6} & \textbf{51.5}\\
            $\Delta$ Cylinder3D & \textbf{\color{fgreen}+5.0} & \textbf{\color{fgreen}+2.2} & \textbf{\color{fgreen}+7.7} & \textbf{\color{fgreen}+6.5} & \textbf{\color{fgreen}+24.8 } & \textbf{\color{fgreen}+12.4}&\textbf{\color{fgreen}+5.9}& \textbf{\color{fgreen}+4.3}& \textbf{\color{bred}-0.5} & \textbf{\color{fgreen}+4.2} & \textbf{\color{fgreen}+4.2} & \textbf{\color{fgreen}+3.3} & \textbf{\color{bred}-0.2} & \textbf{\color{fgreen}+1.2} & \textbf{\color{fgreen}+4.6} & \textbf{\color{fgreen}+1.6} & \textbf{\color{fgreen}+4.9} &\textbf{\color{fgreen}+3.9} & \textbf{\color{fgreen}+2.4} &\textbf{\color{fgreen}+2.8} \\
        \hline
        IGNet++ (Ours) & 63.0 & 94.6 & 44.8 & 67.5 & 78.3 & 55.9 & 72.7 &  85.5 & 0.0  & 88.5 & 42.3 & 75.9 & 2.1 & 90.4 & 53.4 & 87.3 & 70.4 & 70.8 & 63.5 & 52.2 \\
        \hline
    \end{tabular}
    }
    \caption{Weakly-supervised 3D semantic segmentation results on ScribbleKITTI. We not only show results from our proposed image-guidance network (IGNet), but also its performance difference compared to the baseline Cylinder3D model and the results of using IGNet within a class-range-balanced self-training pipeline (IGNet++). * indicated methods that do \textit{not} use Cylinder3D as their backbone.
    \label{tab:results_weakly}}
\end{table*}

\begin{table*}[t]
\begin{minipage}{.57\textwidth}
        \tabcolsep=0.13cm
        \centering
        \begin{tabular}{|l|cccc|cccc|c|}
        \hline
         & \multicolumn{4}{c|}{SemanticKITTI~\cite{iccv2019semantickitti}} & \multicolumn{4}{c|}{ScribbleKITTI~\cite{Unal_2022_CVPR}} \\
        Method & $1\%$ & $10\%$ & $20\%$ & $50\%$ & $1\%$ & $10\%$ & $20\%$ & $50\%$ \\
        \hline
        Cylinder3D~\cite{cvpr2021cylindrical} & 45.4 & 56.1 & 57.8 & 58.7 & 39.2 & 48.0 & 52.1 & 53.8 \\
        \hline
        DiAL~\cite{nips2017meanteacher, unal2023discwise} & 45.4 & 57.1 & 59.2 & 60.0 & 41.0 & 50.1 & 52.8 & 53.9 \\
        CBST~\cite{eccv2018classbalanced} & 48.8 & 58.3 & 59.4 & 59.7 & 41.5 & 50.6 & 53.3 & 54.5 \\
        CPS~\cite{chen2021semi} & 46.7 & 58.7 & 59.6 & 60.5 & 41.4 & 51.8 & 53.9 & 54.8 \\
        GPC~\cite{iccv2021guided} & 34.6 & 49.9 & 58.8 & - & - & -& - & - \\
        WS3D~\cite{liu2022weakly} & 38.9 & 52.3 & 61.4 & - & - & -& - & - \\
        LaserMix~\cite{kong2022lasermix} & \textbf{50.6} & 60.0 & 61.9 & 62.3 & 44.2 & 53.7 & 55.1 & 56.8 \\
        \hline
        IGNet & 49.0 & \textbf{61.3} & \textbf{63.1} & \textbf{64.8} & \textbf{44.4} & \textbf{57.7} & \textbf{59.6} & \textbf{60.8} \\
        $\Delta$ Cylinder3D & \textbf{\color{fgreen}+4.6} & \textbf{\color{fgreen}+5.2} & \textbf{\color{fgreen}+5.3} & \textbf{\color{fgreen}+4.1} & \textbf{\color{fgreen}+5.2} & \textbf{\color{fgreen}+9.7} & \textbf{\color{fgreen}+7.5} & \textbf{\color{fgreen}+7.0} \\
        \hline
        \end{tabular}
        \caption{Comparison of state-of-the-art methods for semi-supervised LiDAR semantic segmentation. The uniform frame sampling rate is indicated by [\%].}
        \label{tab:results_semi}
\end{minipage}
\hfill
\begin{minipage}{.39\textwidth}
    \tabcolsep=0.145cm
    \centering
    \begin{tabular}{|cccc|ccc|}
    \hline
    MT & IG & CL & FOVMix & mIoU & rel & $\Delta$rel \\
    \hline
     & & & & 57.0 & 88.6 & - \\
    \ding{51} & & & & 59.0 & 91.8 & \textbf{\color{fgreen}+3.2} \\
    \ding{51} & \ding{51} & & & 61.3 & 95.3 & \textbf{\color{fgreen}+6.7}\\
    \ding{51} & \ding{51} & \ding{51} & & 61.5 & 95.6 & \textbf{\color{fgreen}+7.0} \\
    \ding{51} & \ding{51} & & \ding{51} & 61.5 & 95.6 & \textbf{\color{fgreen}+7.0} \\
    \ding{51} & \ding{51} & \ding{51} & \ding{51} & 62.0 & 96.4 & \textbf{\color{fgreen}+7.8} \\
    \hline
    \end{tabular}
    \caption{Ablation study where starting from the baseline Cylinder3D, we one-by-one introduce the mean teacher (MT), as well as our proposed image guidance (IG), contrastive loss (CL) and FOVMix modules. Alongside the mIoU, we also report the relative mIoU  (rel) compared to the fully supervised baseline.}
    \label{tab:ablation_components}
\end{minipage}
\vspace{-12px} \end{table*}

\begin{table*}[t]
\begin{minipage}{.48\textwidth}
    \tabcolsep=0.11cm
    \centering
    \begin{tabular}{|cc|cc|cc|}
    \hline
    Source & Target & mIoU & rel & $\Delta$ mIoU & $\Delta$ rel \\
    \hline
    SKITTI (W) & - &  60.3 & 93.7 & - & -\\
    \hline
    GTA-V & - & 60.2 & 93.6 & - & -\\
    GTA-V & SKITTI (U) & 61.1 & 95.0 & \textbf{\color{fgreen}+0.9} & \textbf{\color{fgreen}+1.4} \\
    GTA-V & SKITTI (W) & \textbf{61.3} & \textbf{95.3} & \textbf{\color{fgreen}+1.1} & \textbf{\color{fgreen}+1.7} \\
    \hline
    \end{tabular}
    \caption{Ablation study showing the effects of domain adaptation for the image-guidance network. (U) indicated unsupervised and (W) indicates weakly-supervised training.}
    \label{tab:ablation_da}
\end{minipage}
\hfill
\begin{minipage}{.47\textwidth}
        \tabcolsep=0.125cm
        \centering
        \begin{tabular}{|l|cc|cc|cc|}
            \hline
             & \multicolumn{2}{c|}{Border} &    \multicolumn{2}{c|}{Object} &
            \multicolumn{2}{c|}{Distance} \\
            Method & True & False & Small & Large & 0-25m & 25m+ \\
            \hline
            Cylinder3D & 62.5 & 91.8 & 73.0 & 94.0 & 87.7 & 84.6 \\
            MT & 62.0 & \textbf{92.7} & 76.9 & 95.0 &  88.4 & 85.2 \\
            \hline
            IGNet & \textbf{65.5} & 92.6 & \textbf{83.5} & \textbf{96.5} & \textbf{88.8} & \textbf{87.2} \\
            $\Delta$ MT & \textbf{\color{fgreen}+3.5} & \textbf{\color{gray}-0.1} & \textbf{\color{fgreen}+6.6} & \textbf{\color{gray}+1.5} & \textbf{\color{gray}+0.4} & \textbf{\color{fgreen}+2.0} \\
            \hline
        \end{tabular}
        \caption{Ablation study on ScribbleKITTI showing where the accuracy improves with our proposed image-guidance module.}
        \label{tab:improvements}
\end{minipage}
\end{table*}

\begin{figure*}[t]
    \centering
    \includegraphics[width=\textwidth]{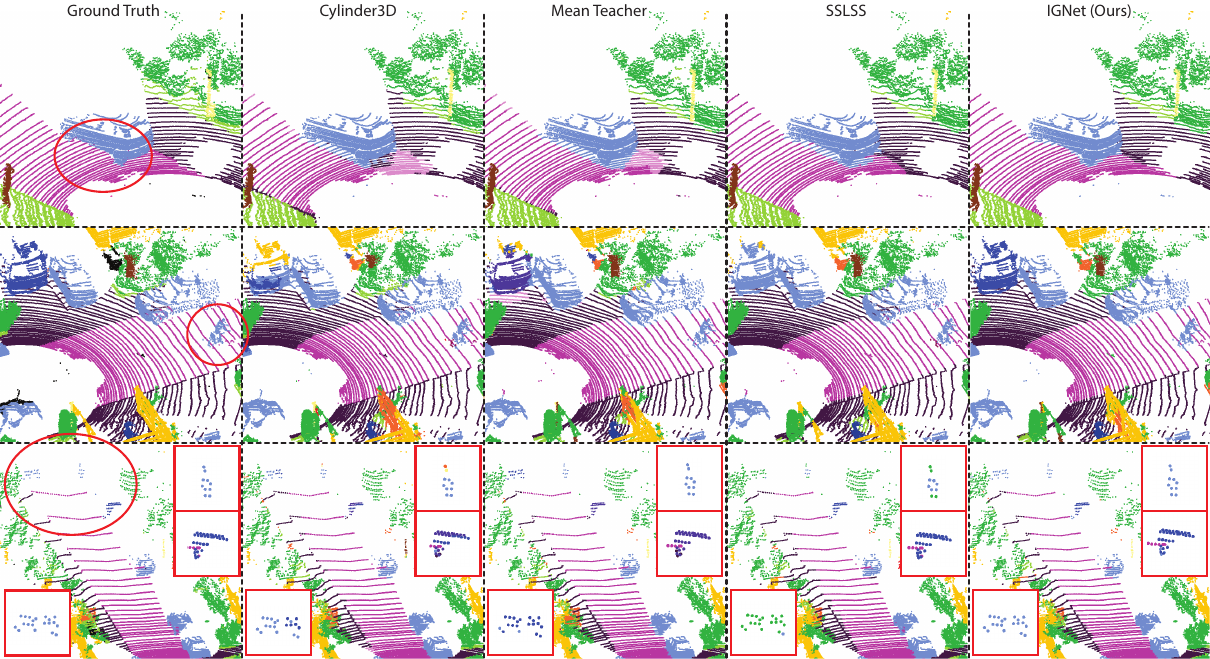}
    \caption{Qualitative results comparing state-of-the-art scribble-supervised LiDAR semantic segmentation methods. As seen, utilizing 2D image features as guidance during the training pipeline, IGNet does improve (top) boundary estimation between classes, (middle) small object segmentation, (bottom) distant, sparse object recognition. We change the color of \textit{bicyclist} in (middle) for better visibility.}
    \label{fig:results}
\vspace{-12px} \end{figure*}

\noindent \textbf{Weakly-Supervised LiDAR Segmentation:} We report the performance of our image-guidance network (IGNet) trained with scribble-supervision in Tab.~\ref{tab:results_weakly}. As seen, IGNet outperforms previous SOTA, showing improvements across the board for all classes and reaching 96.4\% relative performance when compared to fully supervised training while only using $8\%$ labeled points. In specific, we observe large gains for small object categories such as \textit{bicycle} and \textit{motorcycle} when compared to the previous SOTA SSLSS~\cite{Unal_2022_CVPR}.

It should be noted that, in contrast to SSLSS, IGNet does not require self-training. Therefore the training times are considerably reduced (from 5 days to 1 - including the 2D training - using 8 Nvidia RTX2080Ti's). Still, to further push performance, we can IGNet++. Here, we replace the Cylinder3D backbone of SSLSS with IGNet and therefore employ the same class-range-balanced self-training scheme on top of our image guidance to achieve 63\% mIoU, i.e. 98\% relative performance compared to fully supervised.

\noindent \textbf{Semi-Supervised LiDAR Segmentation:} We also show that IGNet can be used for all data-efficient LiDAR semantic segmentation settings. In particular, we report results for (i) semi-supervised training using SemanticKITTI~\cite{iccv2019semantickitti} and (ii) semi- and weakly-supervised training on ScribbleKITTI~\cite{Unal_2022_CVPR}, where we carry experiments on a semi-supervised setting while training with a weakly-supervised dataset. We follow Kong~\etal~\cite{kong2022lasermix} and generate a semi-supervised dataset by uniformly sampling frames.

As seen in Tab.~\ref{tab:results_semi}, IGNet outperforms previous SOTA's by a considerable margin on almost all cases. Specifically, as expected, we see greater margins of improvement in the ScribbleKITTI semi-supervised benchmark since the image-guidance can be more effectively utilized to learn boundary information despite the lack of any such labels. We also report a direct comparison to the baseline Cylinder3D model where IGNet shows great absolute mIoU improvements of $4.1\%-9.7\%$ while introducing no additional memory or computational requirements during inference.

\subsection{Ablation Studies}

We conduct ablation studies on the ScribbleKITTI~\cite{Unal_2022_CVPR} dataset, where alongside the mIoU, we also report the relative performance of our model compared to the baseline Cylinder3D~\cite{cvpr2021cylindrical} trained on densely annotated labels.

\noindent \textbf{Effects of Network Components:} We first investigate the effects our proposed components. Starting from a baseline model, we introduce each module one by one, reporting the mIoU and relative performances in Tab.~\ref{tab:ablation_components}. As seen each component provides a considerable performance gain over the baseline. Specifically we see a $2\%$ gain when we introduce our domain adapted image-guidance network, and a further $0.2\%$ when we introduce our contrastive loss/FOVMix individually. When utilizing both modules, we see that the constrastive loss can benefit from additional point pairings established via the FOVMix operation, which reflects in the gain of $0.8\%$ (as opposed to $0.3\%$).

\noindent \textbf{Is Domain Adaptation Necessary?} We further investigate the necessity of domain adaptation for our image-guidance network. Starting from a mean teacher framework, we compare the performance of our 3D model when guided by the DAFormer model~\cite{hoyer2021daformer} trained on (i) weak labels that we generate by projecting 3D scribbles onto the image, and (ii) the synthetically generated GTA-V dataset~\cite{richter2016playing}, as well as the complete DAFormer pipeline (model + DA) with (iii) GTA-V $\rightarrow$ ScribbleKITTI, and (iv) GTA-V $\rightarrow$ ScribbleKITTI with additional projected weak supervision. The results are shown in Tab.~\ref{tab:ablation_da} which emphasize the importance of DA and the usefulness of the weak supervision.

\noindent \textbf{Where do the Improvements Come From?} Our goal when using image features to guide our 3D model is to exploit the better representation capabilities of 2D semantic segmentation models trained on denser representations for (i) border points, where color channels can provide finer separation compared to noisy LiDAR measurements, (ii) small object and sparsely represented regions, where the pixel count remains considerably higher compared to the LiDAR point count. Finally, we conduct an ablation study to investigate if this behaviour can be observed in the model accuracy after introducing the 2D image-guidance module. 

In Tab.~\ref{tab:improvements}, we isolate the effects of our image guidance module by directly comparing to the mean teacher. Firstly, we show that the introduction of image-guidance does boost the border accuracy significantly ($+3.5\%$). Here, we classify points to be on a border if any of its closes $N=16$ neighbors in 3D space do not share the same class.
Second, we observe that IGNet obtains a considerably better performance ($+6.6\%$) on small objects (pedestrians and two-wheelers) compared to the gain in larger objects ($+1.5\%$ for four-wheelers).
Lastly, when comparing accuracy changes by range, sparsely represented distant regions beyond $25m$ of range show an improvement of $+2.0\%$ when compared to the MT baseline, while close regions only see marginal gains of $+0.4\%$. Here we conclude that image-guidance can indeed compensate for the common weaknesses seen in LiDAR segmentation, especially under weak supervision.

Apart from quantitative results, we also showcase examples from the \textit{valid}-set illustrating this effect in Fig.~\ref{fig:results}. Here we show that IGNet can (top) finely determine object boundaries, (middle) better segment small objects (Cylinder3D and SSLSS misidentify some bicyclist points), and (bottom) improve recognition for sparsely represented regions (IGNet correctly segments all three sparse objects).

\section{Conclusion}

In this work we tackle common weaknesses of data efficient LiDAR semantic segmentation by distilling high level feature information from a synthetically trained 2D semantic segmentation network. We reduce the domain gap between synthetic and real data by employing weakly supervised DA. We extend the supervision from image pixels to out-of-FOV points via a one way contrastive loss and construct new pairings via FOVMix. With our proposed IGNet, we achieve better boundary estimation, increase performance at distant, sparse regions and heavily improve small class segmentation. We achieve SOTA results in both weakly- and semi-supervised 3D semantic segmentation.

\noindent \textbf{Limitations:} Compared to the baseline Cylinder3D, IGNet requires roughly twice the training time due to its two stage approach. Furthermore, the feature distillation module requires paired RGB images with LiDAR scans. While all current LiDAR equipped autonomous systems have an accompanying camera setup, our method still relies on the fact that the sensors need to be calibrated for valid pairings.

\noindent \textbf{Acknowledgements:} This work was funded by Toyota Motor Europe via the research project TRACE Zurich.

{\small
\bibliographystyle{ieee_fullname}
\bibliography{egbib}
}

\clearpage
\clearpage

\end{document}